\documentclass[a4paper,fleqn]{cas-sc}

\usepackage{float}

\usepackage[authoryear,longnamesfirst]{natbib}

\def\tsc#1{\csdef{#1}{\textsc{\lowercase{#1}}\xspace}}
\tsc{WGM}
\tsc{QE}

\begin{document}
\let\WriteBookmarks\relax
\def\floatpagepagefraction{1}
\def\textpagefraction{.001}

\shorttitle{}    

\shortauthors{S. K. Chowdhury et al.}  

\title [mode = title]{Attention-Guided Dual-Stream Learning for Group Engagement Recognition: Fusing Transformer-Encoded Motion Dynamics with Scene Context via Adaptive Gating}  



%

\author[1]{Saniah Kayenat Chowdhury}



\ead{kayenat945@gmail.com}



\affiliation[1]{organization={Department of Robotics and Mechatronics Engineering},
            addressline={University of Dhaka}, 
            city={Dhaka},
            country={Bangladesh}}
\author[2]{Muhammad E. H. Chowdhury}\cormark[1]


\ead{mchowdhury@qu.edu.qa}

\ead[url]{https://orcid.org/0000-0003-0744-8206}


\affiliation[2]{organization={Department of Electrical Engineering},
            addressline={Qatar University}, 
            city={Doha},
            postcode={2713}, 
            country={Qatar}}

\cortext[1]{Corresponding author}



\begin{abstract}
Student engagement is crucial for improving learning outcomes in group activities. Highly engaged students perform better both individually and contribute to overall group success. However, most existing automated engagement recognition methods are designed for online classrooms or estimate engagement at the individual level. Addressing this gap, we propose DualEngage, a novel two-stream framework for group-level engagement recognition from in-classroom videos. It models engagement as a joint function of both individual and group-level behaviors. The primary stream models person-level motion dynamics by detecting and tracking students, extracting dense optical flow with the Recurrent All-Pairs Field Transforms network, encoding temporal motion patterns using a transformer encoder, and finally aggregating per-student representations through attention pooling into a unified representation. The secondary stream captures scene-level spatiotemporal information from the full video clip, leveraging a pretrained three-dimensional Residual Network. The two-stream representations are combined via softmax-gated fusion, which dynamically weights each stream's contribution based on the joint context of both features. DualEngage learns a joint representation of individual actions with overarching group dynamics. We evaluate the proposed approach using five-fold cross-validation on the Classroom Group Engagement Dataset developed by Ocean University of China, achieving an average classification accuracy of $0.9621 \pm 0.0161$ with a macro-averaged F1 of $0.9530 {\pm}0.0204$. To understand the contribution of each branch, we further conduct an ablation study comparing single-stream variants against the two-stream model. This work is among the first in classroom engagement recognition to adopt a dual-stream design that explicitly leverages motion cues as an estimator. 
\end{abstract}




\maketitle

\section{Introduction} \label{sec:intro}
Student engagement is a multifaceted concept that broadly refers to the quality of a student's participation and effort in learning activities \cite{fredricks2004school, christenson2012handbook}. It is primarily categorized into three dimensions: behavioral, emotional, and cognitive. Behavioral engagement involves observable actions such as participation and effort; emotional engagement encompasses students' emotional reactions to tasks; and cognitive engagement reflects their investment in learning through deeper processing and self-regulation. These facets are critical, as they correlate with key academic outcomes, including achievement, persistence, and reduced dropout rates \cite{appleton2006measuring}. 

Over the years, educators have employed various methods to monitor student engagement during classroom activities. Such long-standing approaches include teacher observations, surveys, and self-reports \cite{reeve2006teachers, rafferty2010step,hager2012self}. However, as classroom sizes grow larger and educational environments continue to evolve, these attempts often fall short, providing a limited view of students' participation and emotional investment in the learning process. To address this challenge, the need for automated processes to assess student engagement has become increasingly apparent. Human ratings and manual coding remain highly subjective and cannot scale effectively to accommodate larger student populations \cite{sinatra2015student, wilkinson2012student}. In addition, sensor-based methods are often unstable and raise concerns about deployment, privacy, and ethics. \cite{herlianto2020iot, schneider2015augmenting}. A natural approach is to use pose estimation or skeleton-based representations to model student behavior. However, such methods are sensitive to occlusion, require accurate keypoints, and often miss subtle shifts in posture and motion synchrony \cite{qarbal2024student, lin2021student}. Recent group activity recognition work, a related field to group engagement detection, has made substantial progress in modeling multi-person interactions using transformer and graph-based reasoning, including role-aware transformers and actor–scene relational modeling \cite{pei2023key, jiang2024unveiling}. However, many of these approaches still rely heavily on accurate actor-centric representations and can be sensitive to crowded classroom conditions, partial visibility, and subtle motion patterns that differentiate engagement levels. Further, current engagement recognition models fail to accurately recognize group-level engagement in real classroom environments because they often only focus on individual students, rely on fragile facial or pose cues, and neglect the complex group dynamics that significantly influence engagement. 

Group activity recognition has shown promising results by integrating individual cues, such as pose, motion, and attention, with scene-level structure that extracts spatial layout, shared tempo, and interaction patterns \cite{choi2009they, ibrahim2016hierarchical, yan2018participation}. Moreover, prior work shows that motion descriptors such as optical flow are informative about people's attention shifts, fidgeting, and turn-taking \cite{horn1981determining, sun2010secrets, farneback2003two}. Meanwhile, global encoders capture synchrony and scene context effectively from spatiotemporal data \cite{carreira2017quo, dosovitskiy2020image}. Despite the success of two-stream designs in human action and group activity analysis \cite{xie2019two, azar2018multi, simonyan2014two}, most engagement recognition systems continue to focus either on analyzing single students or narrow areas such as facial expressions, limiting their ability to address complex group dynamics \cite{thomas2017predicting,sharma2022student, whitehill2014faces,gupta2016daisee}. 

This study addresses the challenges of recognizing group-level engagement from classroom videos. We define group engagement as the degree to which group members collectively exhibit observable, on-task participation during the activity interval. Using video clips from the OUC–CGE classroom group engagement dataset \cite{OUC-CGE}, we predict engagement levels categorized into high, low, or medium. Our proposed two-stream framework, DualEngage, consists of (i) a primary, person-level stream and (ii) a secondary, scene-level stream. The primary stream processes individual students by performing detection, tracking, and motion feature extraction. We also employ an attention pooling mechanism that assigns learned importance weights across students to emphasize the most behaviorally salient individuals rather than treating all students equally. The secondary, also called the global stream, extracts spatiotemporal representations from the full video clip via 3D ResNet-18 \cite{ebrahimi2020introducing} backbone. The outputs from both streams provide rich localized representations along with global coordination cues. By combining these outputs via softmax-gated fusion, the model predicts effective student engagement levels by dynamically balancing the contribution of each stream based on their joint context, allowing the model to rely more heavily on scene-level cues when individual motion is ambiguous and on motion cues when individual behavioral patterns are discriminative.

The following are the major contributions that have been presented in our study:
\begin{itemize}
\item Design and implementation of a novel dual-stream architecture for group-level student engagement recognition that models engagement as a joint function of individual motion dynamics and collective scene context.
\item Development of a person-level motion stream that extracts per-student dense optical flow and models its temporal evolution through a shared transformer encoder, capturing long-range behavioral dependencies.
\item Introduction of attention pooling over per-student transformer embeddings to produce an engagement-aware motion representation.
\item Formulation of a scene-level stream using a pretrained 3D ResNet-18 to extract spatiotemporal features from the full video clip, capturing collective dynamics such as group synchrony that cannot be attributed to any single student.
\item Proposal of softmax-gated fusion to adaptively combine the two stream representations, dynamically balancing their contributions based on joint context to compensate when individual motion cues are ambiguous.
\item Comprehensive ablation study validating the contribution of each architectural component, including single-stream baselines, temporal encoding removal, and alternative backbone comparisons.
\end{itemize}

The following sections are structured into four cohesive parts. Section \ref{sec:lit} dives into related work on student engagement, group activity recognition, and the role of motion in determining group collaboration. The methodology is detailed in Section \ref{sec:method}. Experimental results and an in-depth ablation study are presented in Section \ref{sec:experiment}. We offer concluding remarks and limitations with potential future work in Section \ref{sec: conclusion}.

\section{Related Literature}\label{sec:lit}
The detection and analysis of engagement levels in learning activities is paramount for improving learning, retention, and academic performance \cite{academic, retention}. In both traditional classrooms and group activities, high engagement through interactive discussions and immediate feedback significantly enhances learning outcomes and reduces dropout rates and prevents off-task or disruptive behavior spikes \cite{gortazar2024online, bolt2019varying}. This section presents a detailed outline of research done on student engagement and group engagement recognition methods and the potential role of motion for impacting engagement levels. 
\subsection{Student Engagement}
Previous works have employed a wide range of methods, predominantly utilizing computer vision and machine learning, to automatically classify or predict student engagement levels in both traditional and online classroom settings. Due to Covid-19 and technological advancements in general, online learning has gained increasing attention in the last few years \cite{Covidonline}, \cite{covidonline2}, \cite{techonline}. Hence, recent work on recognizing student engagement concentrates overwhelmingly on online learning, where webcams and platform logs are readily available. Whether online or in person, it is imperative to make sure that the students are engaged in the content.

Earlier works using facial Action Units (AUs), head pose, and gaze from OpenFace \cite{baltruvsaitis2016openface} rely heavily on frontal visibility, stable lighting, and clear facial resolution under the premise that these visible landmarks are the primary indicators of a student's attention. \cite{thomas2017predicting}. Thomas et al. \cite{thomas2017predicting} demonstrated that analyzing short video clips for facial and gaze cues could significantly improve classification accuracy over baseline results. Furthering this approach, Sharma et al. \cite{sharma2022student} developed a system utilizing built-in web cameras to extract head movements and facial emotions, resulting in a "concentration index" that categorizes learners into three levels: very engaged, nominally engaged, and not engaged at all. Prior literature also notes that emotion cues alone correlate weakly with cognitive or behavioral engagement during collaborative tasks. As a result, these methods cannot capture the nonverbal interaction patterns that define group engagement \cite{hasnine2021students}. 

Geometry-only pipelines compute inter-landmark angle features from faces; Gopinathan et al. \cite{sherly2024fostering} reported high accuracy  with random forests \cite{rigatti2017random} for low vs. high engagement, indicating lightweight posture cues rival deeper models when framing is stable. Kumar et al. \cite{kumar2025deep} further compared transfer-learned CNNs such as MobileNet \cite{sinha2019thin}, ResNet50 \cite{koonce2021resnet}, VGG16 with MobileNet on a balanced dataset, highlighting its practicality for commodity devices.

As classroom environments are often "in-the-wild" and characterized by occlusions or varied lighting, the limitations of single-modality systems led to the emergence of multimodal fusion frameworks. Monkaresi et al. \cite{monkaresi2016automated} explored the integration of video-based facial expressions with physiological data, such as heart rate estimated via computer vision, to detect engagement during structured writing tasks. In a similar vein, Okur et al. \cite{okur2017behavioral} and Alyuz et al. \cite{alyuz2017unobtrusive} proposed unobtrusive frameworks that complement visual appearance data with contextual logs, such as URL usage and mouse interaction patterns, to distinguish between on-task and off-task behaviors. Synthesizing trends, Dewan et al. \cite{dewan2019engagement} highlighted that combining multiple non-intrusive visual modalities is a practical path to higher accuracy in online environments. Their findings suggest that while facial appearance is a strong primary signal, the inclusion of contextual and performance-based data significantly enhances the robustness of engagement detection in authentic, unstructured learning scenarios.

Recent advancements have also begun to account for the broader classroom ecosystem, including the reciprocal influence of teacher behavior. Verma et al. \cite{verma2023designing} developed a tool to evaluate student engagement by analyzing the movements and behaviors of instructors during video conferencing, suggesting that engagement is not an isolated student state but a social process influenced by teaching quality. In the context of online learning, Hasnine et al. \cite{hasnine2021students} focused on extracting and visualizing emotions from lecture videos to provide teachers with actionable feedback, thereby bridging the physical gap between instructor and student. However, despite these innovations, a significant research gap persists in the analysis of group-level engagement within physical classrooms. Most existing models are designed for individual students or online environments and fail to capture the collective synergy and social interactions that define group activities. Group engagement necessitates architectures that can process both local and global dependencies simultaneously.

\subsection{Group Engagement and Activity}
Modeling group-level engagement in learning contexts remains a relatively unexplored domain and is largely online or video-centric. Pabba et al. \cite{pabba2022intelligent} present a real-time classroom monitor that detects faces, recognizes “academic affective states,” and aggregates frame-wise predictions into a class-level engagement estimate—demonstrating scalable but vision-only pooling suited to fixed cameras. However, group activity has been well explored in areas beyond education for action or group recognition. Frank et al. \cite{frank2016engagement} propose a multi-modal engagement framework for meetings combining vision and audio, estimating individual states and aggregating to team state via modular classifiers. Given the scarcity of classroom-specific group engagement studies, methods from general group activity recognition are instructive. Pose and skeleton-based approaches have also been widely used for multi-person behavior understanding, representing each person as a set of joints and modeling dynamics through spatio-temporal graphs \cite{yan2018spatial,lovanshi2024human}. In these methods, joints are treated as graph nodes, and temporal edges connect joints across frames, enabling action reasoning with reduced dependence on appearance. In classroom settings, such methods can be effective when reliable key points are available, but performance may degrade under heavy occlusion, truncation, and crowded group layouts where joints are partially missing or noisy. Among multi-stream approaches, stagNET \cite{qi2019stagnet}  models scenes as semantic spatio-temporal graphs with attention to infer both person-level actions and the group activity, explicitly capturing inter-person dependencies that drive collective behavior. This two-stream approach can be highly utilized in recognizing group engagement level, where individual engagement impacts group engagement.  Matz-Costa et al. \cite{matz2019perceptions} examine how different levels of productive activity engagement, such as paid work, volunteering, and caregiving, relate to psychological well-being in midlife and later life. Using large cohort data, they show that greater or more varied role engagement is associated with better well-being, framing engagement as a measurable construct with public health relevance. More recently, transformer-based multi-person action and group activity recognition methods have been explored that use attention to model interactions among multiple individuals while learning global context \cite{shabaninia2022transformers,wang2025multi}. These approaches provide strong multi-person baselines by explicitly modeling inter-person relations. However, many still rely on stable, person-centric inputs. Our approach is complementary, as we emphasize dense motion to capture subtle engagement-linked movement patterns and fuse these with a scene-level spatiotemporal encoder to obtain global context.

\subsection{Motion}
A growing body of work suggests that overt movement dynamics can serve as informative proxies for engagement, with evidence spanning gameplay, multimodal sensing, and group-activity analysis. In a controlled manipulation, Leiker et al. \cite{leiker2016relationship} altered graphical richness in a kinect-based game to form an explicit link between motion-coupled interaction and cognitive resources relevant to learning. Complementing this, Lindley et al. \cite{lindley2008stirring} contrasted motion-affording bongos with a standard controller to examine whether increased real-world social behavior attenuates game-world immersion, directly testing how input-elicited movement shapes engagement. Similarly, Bethouze et al. \cite{bianchi2007does} advanced the premise that greater body involvement—imposed or allowed by the controller—can intensify affective engagement, supported by two case studies.

Beyond motion as a single modality, Huynh et al. \cite{huynh2018engagemon} integrated skeletal kinematics with touch across six mobile games, using expert feedback to craft features and models for continuous engagement estimation. This proves that motion features synergize with physiological and interaction signals. Extending to the collective level, Stephens et al. \cite{stephens2016human} identified group activity by modeling interdependencies among motion flows and locations using Kernel Density Estimation (KDE) in joint time–space and time–movement spaces, enabling classification without manual tracking. In group settings, motion provides synchrony cues, such as coordinated attention shifts, leaning, or collective orientation, that facial analysis cannot capture, particularly when faces are partially occluded or oriented away from the camera.

\section{Methodology}\label{sec:method}
\begin{figure}[htbp]
    \centering
    \includegraphics[width=1\linewidth]{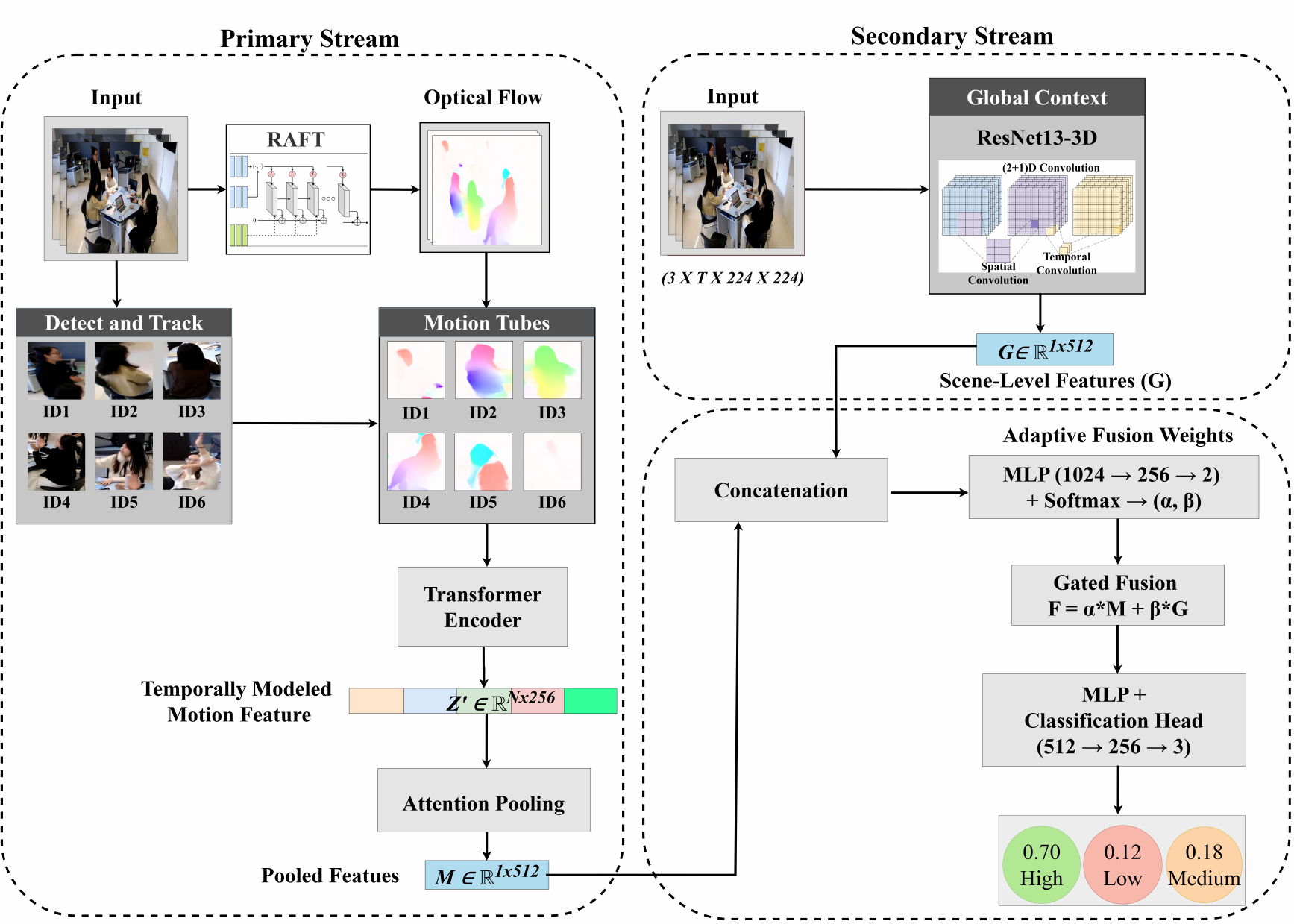}
    \caption{Overview of the proposed DualEngage. In the primary stream, optical flow is computed using RAFT; per-student motion tubes are cropped using tracked identities and modeled with a transformer encoder to yield a unified motion embedding before being aggregated via attention pooling. In the secondary stream, a 3D ResNet extracts scene-level features from the full spatiotemporal clip. The two feature sets are combined using gated fusion and passed to an MLP classification head to predict overall engagement levels during group activities.}
    \label{fig:methodology}
\end{figure}

The proposed methodology in this work introduces a two-stream architecture, named DualEngage, for group activity student engagement recognition (Figure \ref{fig:methodology}). As suggested by the name, the architecture comprises two distinct branches: 
\begin{itemize}
    \item A Primary stream, also denoted as person-level branch, focused on analyzing engagement at the individual level.
    \item A secondary or scene-level stream providing a global context derived from the classroom videos. 
\end{itemize}
In the primary stream, we emphasize person-level motion dynamics. This sub-architecture detects and tracks students, generating individual student clips from the video using libraries from OpenMMLab \cite{chen2019mmdetection, wojke2017deepsort}. These clips are processed through the Recurrent All-Pairs Field Transforms (RAFT) \cite{teed2020raft} optical flow model \cite{horn1981determining}, which creates motion tubes that capture the temporal evolution of individual student movements. We use RAFT for optical flow estimation because its all-pairs correlation volume and iterative refinement enable it to capture subtle, non-rigid motions that are often missed by lighter models such as PWC-Net \cite{sun2018pwc}. To preserve their temporal structure, a shared transformer encoder \cite{vaswani2017attention} is employed. It uses a single encoder to process the entire sequence of motion data simultaneously, utilizing positional encoding to capture the temporal order of movements. The output is a set of individualized features corresponding to each student's engagement. To understand behaviorally salient individuals, the resulting embeddings or motion feature set is subsequently collapsed into a single vector via attention pooling. A linear projection scores 
each student, and a weighted sum normalized using softmax produces the final motion 
representation. The secondary stream considers the full context of the group by extracting information from interactions between students and capturing the overall group flow. For this, we leverage a pre-trained 3D ResNet-18 \cite{ebrahimi2020introducing}, which extracts complex temporal and spatial features from classroom videos. At the fusion stage, 
instead of simple concatenation, we implement an adaptive gated fusion, described in \ref{attn_pool}. The resulting representation is passed to a classification head to predict one of three engagement levels: high, low, or medium.

\subsection{Data Acquisition and Pre-processing}
In this section, the details of the data used for the analysis are described. The procedure of data processing is also discussed in detail. The OUC-CGE \cite{OUC-CGE} dataset is the first benchmark dedicated to group engagement analysis using only visual signals. This dataset was constructed in real classrooms instrumented with three C922 Pro webcams recording in 1280×720 resolution at 30 fps. These cameras were arranged at three angles to capture group activities from diverse perspectives, and there were two room layouts: round-table and chessboard for variability. The recorded videos were segmented into approximately 10 seconds clips annotated at three engagement levels: low, medium, and high. The annotations are done through a dual-coding protocol aligned with Interactive, Conductive, Active, Passive (ICAP) protocol-based observational criteria. Although engagement can be studied as behavioral, emotional, or cognitive engagement, the labels in the dataset correspond primarily to externally observable behavioral engagement. Sustained on-task attention and frequent participation or interaction cues across multiple members are interpreted as high engagement, medium as intermittent attention with occasional off-task behaviors, and low as predominant off-task orientation and minimal participation cues. Our method therefore targets behavioral engagement recognition rather than self-reported cognitive engagement or explicit emotion recognition. The data descriptor reports a total of 7,705 labeled video segments, obtained from 21 sessions with groups of 7–9 students and multiple camera viewpoints, and motivates the 10 second window as a balance between temporal granularity and behavioral coherence. Upon analyzing the public release, we found that the number of available clips did not match the figure reported in the descriptor. Concretely, the dataset in actuality contained 6,700 clips. This discrepancy likely reflects subsequent curation of the online repository, such as removals for privacy or quality relative to the paper’s static count. We therefore treat the descriptor’s statistics as authoritative while reporting all training and evaluation numbers with respect to the actually retrievable files. A metadata pass over the downloaded videos revealed class-specific variability in native frame rates and temporal spans detailed in table \ref{tab:dataset}.
\begin{table}[t]
\centering
\caption{Dataset Summary (Before Pre-processing)}
\label{tab:dataset}
\begin{tabular}{llll}
\toprule
Engagement Level & Number of Clips & Frame Rates (fps) & Duration (s)  \\
\midrule
High & 2,648 &  22-30 & 5-15  \\
Medium & 1,012 &  15-30 & 5-15  \\
Low & 3,040 &  2-30 & 2-15  \\
\bottomrule
\end{tabular}
\\[0.2em]
\end{table}
These ranges are consistent with the recording pipeline described in the dataset paper, but exhibit practical deviations due to encoding and trimming. To eliminate model confounds induced by heterogeneous frame rates, we temporally resample all clips to a uniform 30 fps using Fast Forward Moving Picture Experts Group (FFmpeg)’s frame duplication and interpolation \cite{newmarch2017ffmpeg}. For clips longer than 10 seconds, we center-crop them to 10, and for shorter clips, we loop the video from the start, repeating frames in order until the total duration reaches 10 seconds. This follows the dataset’s intended 10 second video clip window while ensuring a fixed temporal input length for training. Frames are decoded to RGB and resized to $(H_0\times W_0)$ = $(224\times 224)$ pixels with bilinear interpolation, preserving aspect ratio via letterboxing when necessary. Pixel intensities are normalized using ImageNet mean and variance \cite{deng2009imagenet}. Furthermore, we have discarded faulty videos that we could identify by any of the following criteria: (i) unreadable container/codec, (ii) fewer than 2 valid frames after decode, or (iii) severe timestamp discontinuities. After this quality-control step, the remaining class counts and other properties are presented in table \ref{tab:datasetpost}.
\begin{table}[t]
\centering
\caption{Dataset summary (After Pre-processing)}
\label{tab:datasetpost}
\begin{tabular}{llll}
\toprule
Engagement Level & Number of Clips & Frame Rates (fps) & Duration (s)  \\
\midrule
High & 2,070 &  30 & 10  \\
Medium & 1,012 & 30 & 10  \\
Low & 2,834 &  30 & 10  \\
\bottomrule
\end{tabular}
\\[0.2em]
\end{table}

\subsection{Detecting and Tracking Individuals} \label{sec:detTrack}

\begin{figure}[htbp]
    \centering
    \includegraphics[width=1\linewidth]{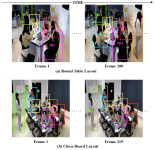}
    \caption{Frame-by-frame student detection and tracking in classroom video. Top (round-table layout): a student enters at frame 208, and is assigned a new ID. Bottom (chessboard layout): two students are intermittently missed being detected due to congestion and occlusion. Colored boxes denote track IDs across frames.}
    \label{fig:DetTrack}
\end{figure}

Capturing motion cues of each individual requires identifying students separately. For this, the students need to be detected and tracked with IDs to have their motion features computed frame by frame. Therefore, we adopt a detection \( \to  \) tracking pipeline using OpenMMLab's libraries. OpenMMLab provides state-of-the-art results for both object detection and tracking tasks, making it an ideal choice for our purpose. The detection is performed using the MMDetection \cite{mmdetection2019} library at each frame of the video. To track students as separate entities, we utilize MMTrack \cite{wojke2017deepsort}, an extension of MMDetect's capabilities. For each detection, the tracker is instantiated to track that person through the next few seconds of video, illustrated in figure \ref{fig:DetTrack}.
\subsubsection{Detect}

Specifically, the detection is performed using Faster R-CNN \cite{ren2016faster} with ResNet-50 \cite{koonce2021resnet} and Feature Pyramid Network (FPN) \cite{lin2017feature} as backbones. This is a two-stage detector implemented within the MMDetection framework. The input to the detection model is a video frame \( I \in \mathbb{R}^{H_0 \times W_0 \times 3} \), where \( H_0 \) and \( W_0 \) represent the height and width of the frame resized to $224\times224$, and the 3 corresponds to the RGB color channels. The Faster R-CNN model outputs a set of bounding boxes \( B = \{b_1, b_2, \dots, b_n\} \), where each bounding box \( b_i = (x_1, y_1, x_2, y_2) \) denotes the coordinates of the detected region, and each detection is assigned a confidence score \( s_i \). To filter the most reliable detections, we use a score threshold \( \tau = 0.8 \) for the person class, ensuring that only high-confidence detections are passed on to the tracking stage.

\subsubsection{Track}
Across frames, we associate detections into trajectories with Deep SORT \cite{wojke2017deepsort}, which augments the original SORT framework’s constant-velocity Kalman filter and IoU gating \cite{bewley2016sort} with a deep appearance metric. The resulting cost combines (i) a Mahalanobis distance between predicted and observed box states relating to motion consistency and (ii) a cosine distance in an appearance-embedding space indicating visual consistency. The 4D state space \( x_t = (x, y, \dot{x}, \dot{y}) \) represents the position and the velocity of the frame. Additionally, we configure two control parameters to reflect the social dynamics of the classroom. The first one is \textit{n\_init} = 3, which indicates that a detection needs to be observed in three consecutive frames before it becomes a confirmed track, filtering transient false positives. The second controlled parameter is \textit{max\_age} = 300, which allows a track to persist for approximately 10 seconds at 30 FPS without observation, tolerating momentary occlusions or brief detector failures while maintaining a consistent track ID. The tracking results provide track IDs and bounding box coordinates, ready to be processed to generate motion tubes.

\subsection{Motion Cues}
We compute dense optical flow for the entire video using the principles of optical flow. Alternatively, pose features could have been highly effective if accurate keypoints were available, but in real group scenes they degrade under occlusion, crowding, camera distance, and truncation. Since the setting in the dataset samples frequently includes partial bodies, overlapping participants, and viewpoint variation, pose estimation can become unstable and may miss subtle motion cues. Optical flow remains computable under partial visibility and captures motion even when keypoints are unreliable, making it a suitable choice for our work. Optical flow models the apparent motion of image intensities and is commonly introduced via the optical flow constraint,
\begin{equation}
I_x u + I_y v + I_t = 0
\end{equation}
\noindent where $I_x, I_y, I_t$ denote spatiotemporal derivatives. Rather than explicitly solving this constraint, we estimate $\mathbf{F}_t$ using the RAFT network \cite{teed2020raft}  pretrained pretrained on FlyingChairs \cite{Dosovitskiy2015FlowNet} and FlyingThings3D \cite{Mayer2016SceneFlow}, then fine-tuned on a Sintel mix comprising Sintel \cite{Butler2012Sintel}, KITTI Flow \cite{Kondermann2016HD1K}, HD1K \cite{Kondermann2016HD1K}, and Things-Clean \cite{Mayer2016SceneFlow}. We run the model on classroom videos in evaluation mode for all consecutive frames. RAFT’s all-pairs correlation and iterative refinement preserve small motions and sharp motion boundaries in classroom scenes.  The videos contain $T$ RGB frames $\{I_t\}_{t=1}^T$, each of size $H_0\times W_0\times 3$. For optical-flow inference, each frame is bilinearly resized to
$\tilde I_t\in\mathbb{R}^{128\times128\times 3}$. This yields, for each adjacent pair $(\tilde I_t,\tilde I_{t+1})$, $t=1,\dots,T-1$, a dense flow map with two channels $(u,v)$ 
\[
\mathbf{F}_t(x,y)=\big[u_t(x,y),\,v_t(x,y)\big]\in\mathbb{R}^{2},
\qquad \mathbf{F}_t\in\mathbb{R}^{2\times128\times128},
\]
where $(u_t,v_t)$ are horizontal and vertical displacements at locations $(x,y)$ on the $128\times128$ grid, indicating how far that pixel moved between two frame.
Stacking over time yields the per-video flow tensor
\begin{equation}
\mathbf{F}=\big[\mathbf{F}_1,\ldots,\mathbf{F}_{T-1}\big]\in\mathbb{R}^{(T-1)\times 2\times128\times128}.
\end{equation}
From section \ref{sec:detTrack}, detection and tracking on the original frames produce, for each person $j$, per-frame bounding boxes
$B_t^{(j)}=(x^{(j)}_{1,t},y^{(j)}_{1,t},x^{(j)}_{2,t},y^{(j)}_{2,t})$ at time $t$ and a stable track ID =$j$.
Since detection runs at the original size $H_0\times W_0$, we map boxes to the RAFT grid via
\begin{equation}
s_x=\frac{128}{W_0},\qquad s_y=\frac{128}{H_0},\qquad
\tilde B_t^{(j)}=\big(s_x x^{(j)}_{1,t},\,s_y y^{(j)}_{1,t},\,s_x x^{(j)}_{2,t},\,s_y y^{(j)}_{2,t}\big).
\end{equation}
Thus $\tilde B_t^{(j)}$ is aligned with the $128\times128$ flow domain.

\noindent For each track $j$ and each $t=1,\dots, T^{(j)}-1$ where the track is present, we crop the full-frame flow at the mapped box:
\begin{equation}
\mathbf{C}_t^{(j)}=\mathrm{Crop}\big(\mathbf{F}_t;\,\tilde B_t^{(j)}\big)\in\mathbb{R}^{2\times h_t^{(j)}\times w_t^{(j)}}.
\end{equation}
To standardize spatial size we bilinearly resample each crop to a fixed tube resolution of $128\times128$
\begin{equation}
\hat{\mathbf{C}}_t^{(j)}=\mathrm{Resize}\big(\mathbf{C}_t^{(j)}\rightarrow 128\times128\big)\in\mathbb{R}^{2\times128\times128}.
\end{equation}
Stacking these over time gives the motion tube for student $j$
\begin{equation}
\mathbf{F}^{(j)}=\big[\hat{\mathbf{C}}_1^{(j)},\ldots,\hat{\mathbf{C}}_{T^{(j)}-1}^{(j)}\big]
\in\mathbb{R}^{(T^{(j)}-1)\times 2\times128\times128}.
\label{eq:motion_tubee}
\end{equation}
Finally, for $N$ number of students, we obtain our motion sequences by stacking motion tubes of each student
\begin{equation}
\mathbf{Z}=\big[{\mathbf{F}}^{(1)},\ldots{\mathbf{F}}^{(N)}\big]
\in\mathbb{R}^{N\times (T-1)\times 2\times128\times128}.
\label{eq:motion_tube}
\end{equation}
We save each $\mathbf{Z}$ as \textit{.npy} for downstream models.
\begin{figure}[htbp]
    \centering
    \includegraphics[width=1\linewidth]{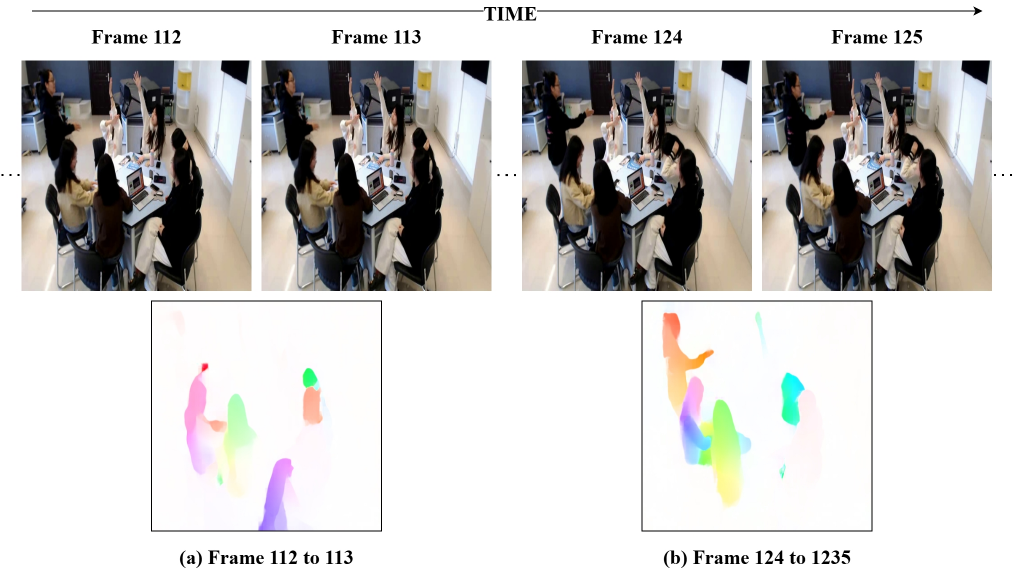}
    \caption{Optical flow map between two consecutive frames.}
    \label{fig:Motion}
\end{figure}

\subsubsection{Temporal Modeling}
\begin{figure}[htbp]
    \centering
    \includegraphics[width=1\linewidth]{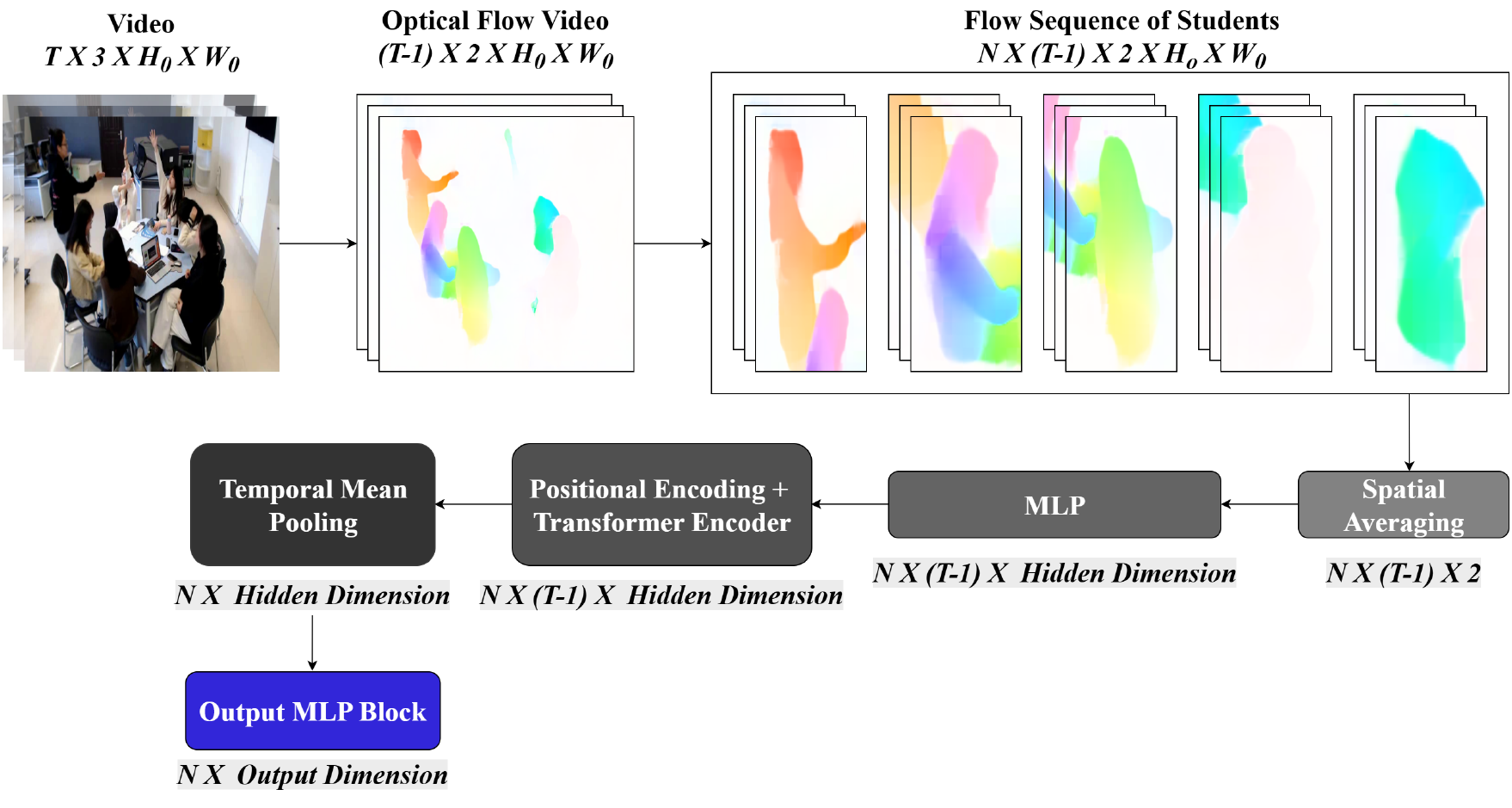}
    \caption{Overview of the motion extraction process from classroom video to student-specific sequences. The raw video is first converted into an optical flow map, followed by the formation of per-student flow sequences. Temporal dynamics are learned using a transformer encoder, which produces the stacked feature representation.}
    \label{fig:motion}
\end{figure}

After extracting the individual motion features ${F}^{(j)}$ and stacking them together in $\mathbf{Z}$, we obtain a summary of movement characteristics and proceed to model the long-range dependencies inherent in the motion data. As engagement behaviors in classrooms often evolve gradually, the motion features $\mathbf{Z}$ are sequential in nature and capture temporal patterns across frames. To effectively capture these dependencies, we utilize a transformer encoder \cite{vaswani2017attention}, which has demonstrated significant success in modeling sequential data by leveraging its self-attention mechanism in prior works \cite{wu2020deep,chen2023contiformer}. We chose this over Recurrent Neural Networks (RNNs) \cite{hochreiter1997long} or Temporal Convolutional Networks (TCNNs) \cite{chen2021deep} due to the transformer's superior capacity for modeling long-range temporal dependencies. Engagement behaviors often unfold slowly, in gradual posture changes and sustained attention shifts. While RNNs suffer from information decay over long sequences, and TCNNs rely on a rigid, fixed receptive field structure, the transformer's self-attention mechanism directly models the relationship between all distant time steps with a constant path length, making it uniquely suited to capture these non-local behavioral patterns \cite{tay2020long}. The transformer encoder is used to process the sequence of motion features over time. Each motion tube $\mathbf{Z}$ is treated as a sequence of tokens, where each token corresponds to the flow features of one student across multiple frames.  These flow features are passed as input to the transformer encoder, which applies the self-attention mechanism to learn contextual relationships between the motion patterns of different students across time.
The output of the encoder is then directly used for the fusion of the two streams. The complete process of motion cue extraction is detailed in figure \ref{fig:motion}, along with dimensions of the features at each step.

\paragraph{Token formation.}
For student $j$, we spatially average the flow crop to obtain
$m^{(j)}_t\in\mathbb{R}^{2}$, which is then projected to the Transformer token dimension via an MLP.
A learnable positional embedding is added to preserve temporal order:
\begin{equation}
x^{(j)}_t=\phi\!\left(W\,m^{(j)}_t+b\right)+p_t,\qquad x^{(j)}_t\in\mathbb{R}^{d},
\label{eq:motion_token}
\end{equation}
where $\phi(\cdot)$ is ReLU, $p_t\in\mathbb{R}^{d}$ is the positional embedding, and $d=512$.

\paragraph{Self-attention.}
Given the token sequence $X^{(j)}=[x^{(j)}_1,\dots,x^{(j)}_{T-1}]$, each Transformer layer applies
scaled dot-product self-attention:
\begin{equation}
\mathrm{Att}(Q,K,V)=\mathrm{softmax}\!\left(\frac{QK^\top}{\sqrt{d_h}}\right)V,
\label{eq:self_attention}
\end{equation}
thereby enabling direct interactions between distant time steps, where $d_h=d/H$ is the per-head
dimension.

\paragraph{Hyperparameters.}
We use a Transformer encoder with $L=3$ layers, $H=8$ heads, feedforward dimension $1024$, and
dropout $0.1$.
The encoder outputs are temporally pooled to obtain one embedding per
student, followed by an output MLP. Collecting the
per-student embeddings yields the final temporally encoded motion representation:
\begin{equation}
Z'=\big[z'^{(1)},\dots,z'^{(N)}\big]\in\mathbb{R}^{N\times 256},
\label{eq:Zprime}
\end{equation}
which is used for subsequent fusion with the global context stream.

\subsubsection{Attention Pooling} \label{attn_pool}
Following temporal modeling, the transformer encoder produces one embedding per student, 
yielding the representation $Z' = [z'^{(1)}, \ldots, z'^{(N)}] \in \mathbb{R}^{N \times 256}$. 
A naive aggregation of these embeddings via mean pooling treats all students equally, which is 
an unrealistic assumption in group settings where engagement levels naturally vary across 
individuals. To address this, we aggregate the per-student embeddings using attention pooling, 
which assigns a learned importance score to each student.

Specifically, a linear projection maps each student embedding $z'^{(i)} \in \mathbb{R}^{256}$ 
to a scalar score, which is normalized across all $N$ students via softmax to produce attention 
weights:

\begin{equation}
    \tilde{a}_i = \frac{\exp(\mathbf{w}^\top z'^{(i)})}
    {\sum_{k=1}^{N} \exp(\mathbf{w}^\top z'^{(k)})}
\end{equation}

The attended motion representation is then computed as the attention-weighted sum and passed through a two-layer MLP with ReLU activations to project it to a higher-dimensional space, yielding the final motion representation
\begin{equation}
    M = MLP\left(\sum_{i=1}^{N} \tilde{a}_i \cdot z'^{(i)} \in \mathbb{R}^{512} \right)
\end{equation}

Here, $\mathbf{w} \in \mathbb{R}^{512}$ is a learnable projection vector. This formulation 
follows the linear attention pooling operator, a simplified variant of the 
attention-based deep multiple instance learning introduced by Ilse et al.~\cite{ilse2018attention}.

\subsection{Global Context Encoder}
With temporal modeling, we complete person-level feature extraction and focus on the secondary stream of DualEngage, which is deployed to summarize the dynamics of the whole scene that cannot be attributed to a single student. These dynamics include class synchrony, teacher pacing, shared interaction shifts, or collective postural changes. Unlike the motion stream, which only monitors individual movements, the global encoder operates on the full clip and produces an embedding that captures scene layout and temporal evolution. An extractor is required to model spatiotemporal receptive fields that jointly integrate motion and appearance over the entire classroom view. 3D CNNs provide such inductive bias by extending 2D kernels with a temporal dimension and producing usable features under occlusions and camera artifacts \cite{yu20202d, hara2018can, carreira2017i3d}. We instantiate this scene-level stream with a 3D ResNet-18 (R3D) \cite{ebrahimi2020introducing} backbone pretrained on Kinetics \cite{kay2017kinetics}. 3D residual blocks perform local spatiotemporal filtering, while skip connections stabilize optimization and enlarge the effective temporal receptive field \cite{he2016resnet, hara2018can}. Compared with deeper 3D CNNs, ResNet-18 offers a favorable trade-off between accuracy and efficiency, especially for a heavy two-stream architecture. Following standard practice in video recognition, we apply Global Spatiotemporal Average Pooling (GAP) to project the feature volume to a single vector that is invariant to small spatial misalignments and clip length, which empirically improves generalization \cite{lin2013network, carreira2017i3d}. A lightweight projection head further maps the pooled vector to the model’s fusion dimension. Our implementation wraps the final pooling and two fully connected layers around the pretrained backbone to yield a compact video descriptor.  Each preprocessed RGB clip of $T$ frames ($224{\times}224$) is stacked as a tensor $(3\times T\times H_0\times W_0)$ and normalized with ImageNet \cite{deng2009imagenet} statistics to match the backbone’s pretraining domain. Temporal strides and pooling within the backbone subsample the sequence while preserving sufficient temporal resolution for classroom events. The resulting embedding encodes static context, meso-scale motion, and interactions with instructional artifacts, all of which correlate with engagement level. The global stream produces a fixed-dimensional vector $\mathbf{G}\in\mathbb{R}^{512}$ per clip, used downstream as one half of the dual stream fusion. By design, $\mathbf{G}$ is invariant to the number of students, complementing the primary branch's motion embeddings.

\subsection{Group Activity Engagement Classification}

The DualEngage architecture classifies group activity engagement by combining the 
temporally modeled motion representation and the global scene context. We obtain a unified motion representation 
$M \in \mathbb{R}^{512}$ from the primary motion stream. The secondary stream employs a pretrained 3D ResNet-18 
to extract spatiotemporal features from the full video clip, producing a global 
scene embedding $G \in \mathbb{R}^{512}$.

Simple concatenation of $M$ and $G$ implicitly assumes equal contribution from 
both streams regardless of the input, which is an unrealistic assumption in 
classroom settings where the discriminative power of each stream varies by clip. 
For instance, in low-motion clips where students remain physically still, 
individual motion cues become less informative and the model should rely more 
heavily on global scene context. To address this, we propose softmax-gated fusion 
to adaptively combine the two stream representations on a per-sample basis.

Specifically, $M$ and $G$ are concatenated and passed through a lightweight MLP to produce two scalar fusion weights:

\begin{equation}
    [\alpha, \beta] = \text{softmax}\left(\text{MLP}([M \| G])\right) \in \mathbb{R}^{2}
\end{equation}

The fused representation is then computed as a convex combination of the two 
stream features:

\begin{equation}
    h = \alpha \cdot M + \beta \cdot G \in \mathbb{R}^{512}
\end{equation}

Since $\alpha + \beta = 1$ by construction, the gate dynamically balances the 
two streams on a per-sample basis without introducing additional constraints 
or regularization. When individual motion cues are ambiguous, the model learns 
to assign a higher weight $\beta$ to the global scene context; conversely, 
when individual behavioral patterns are discriminative, a higher weight $\alpha$ 
is assigned to the motion representation. The final engagement logits are 
computed as:

\begin{equation}
    \ell = \mathbf{W}h + b \in \mathbb{R}^{3}
\end{equation}

where $\mathbf{W}$ is the weight matrix and $b$ is the bias term of the 
classification head, producing scores over the three engagement levels: 
high, low, and medium.

\section{Experiment and Results}\label{sec:experiment}
This section details the experimental framework. We describe the setup for each stream of DualEngage and evaluate performance on group engagement level prediction using significant classification metrics, including accuracy, precision, recall, and macro-averaged F1. We further present a comprehensive ablation study that isolates the contributions of the primary motion stream, 
the secondary scene stream, and the fusion strategy, examining multiple backbone variants to validate the effectiveness of the two-stream design.

\subsection{Experimental Setup}
To achieve the best performance, training parameters were fine-tuned in the classification stage. Experiments were conducted on a local workstation equipped with an AMD EPYC 74F3 processor (18 cores, 36 threads) and 433 GB of RAM, running a 64-bit x86\_64 architecture on Ubuntu 22.04.5 LTS. GPU acceleration was provided by an NVIDIA A10 with 24 GB of VRAM. The software environment comprised Python 3.10.6 and PyTorch 2.6.0{+}cu118.
We evaluated DualEngage using stratified 5-fold cross-validation. The model was trained with the Adam optimizer \cite{kingma2015adam} at a learning rate of $1{\times}10^{-4}$ and cross-entropy loss with inverse-frequency class weights to mitigate class imbalance. A ReduceLROnPlateau scheduler (factor $0.5$, patience $10$, $\text{min\_lr}=1{\times}10^{-7}$) adjusted the learning rate on validation plateaus. Early stopping monitored validation F1 (patience $20$, $\text{min\_delta}=1{\times}10^{-4}$); the best checkpoint was selected by higher validation F1, tie-broken by lower validation loss. Hyperparameter details are summarized in Table~\ref{tab:params}.
\begin{table}[t]
\centering
\caption{Training Setup}
\label{tab:params}
\begin{tabular}{ll}
\toprule
Parameter & Value \\
\midrule
Input clip length & $T{=}300$  \\
Frame Size (scene-level stream) & $224{\times}224$ \\
Frame Size (person-level stream) & $128{\times}128$ \\
Batch size / workers & $16$ / $4$ \\
Optimizer / LR & Adam / $1\!\times\!10^{-4}$ \\
Loss & Cross-entropy with inverse-frequency class weights \\
Scheduler & ReduceLROnPlateau  \\
Early Stopping &  Patience $20$  \\
Folds & $5$ \\
Epoch  & $100$  \\
\bottomrule
\end{tabular}
\end{table}
\subsection{Evaluation Metrics}
At the end of each fold, we evaluate on that fold's test split across three classes. Following the training script, we report overall accuracy and macro-averaged precision, recall, and F1 and summarize results as mean~$\pm$~standard deviation across folds. Macro-F1 is informative under class imbalance because it weights each class equally, preventing the majority class from dominating the overall score. We also provide confusion matrices highlighting the per-class true positives, false positives, true negatives, and false negatives. Per-class precision, recall, and F1-score are defined as follows:
\begin{align}
\mathrm{Prec}_c &= \frac{TP_c}{TP_c+FP_c+\varepsilon},\\
\mathrm{Rec}_c  &= \frac{TP_c}{TP_c+FN_c+\varepsilon},\\
\mathrm{F1}_c   &= \frac{2\,\mathrm{Prec}_c\,\mathrm{Rec}_c}{\mathrm{Prec}_c+\mathrm{Rec}_c+\varepsilon},
\end{align}
with a small $\varepsilon>0$ to guard against division by zero. We use \emph{macro-averaging} to
account for class imbalance by assigning equal weight to each class:
\begin{align}
\mathrm{Precision}_{\mathrm{macro}} &= \frac{1}{|\mathcal{C}|}\sum_{c\in\mathcal{C}} \mathrm{Prec}_c,\\
\mathrm{Recall}_{\mathrm{macro}}    &= \frac{1}{|\mathcal{C}|}\sum_{c\in\mathcal{C}} \mathrm{Rec}_c,\\
\mathrm{F1}_{\mathrm{macro}}        &= \frac{1}{|\mathcal{C}|}\sum_{c\in\mathcal{C}} \mathrm{F1}_c.
\end{align}
Overall accuracy is
\begin{equation}
\mathrm{Accuracy} \;=\; \frac{1}{N}\sum_{c\in\mathcal{C}} TP_c \;=\; \frac{1}{N}\sum_{i=1}^{K} M_{ii}.
\end{equation}
This prevents the majority class from dominating the overall score and better reflects how well the model performs on minority classes that are often harder to recognize in classroom videos.

\subsection{Results and Discussion}  
\begin{figure}[htbp]
  \centering
   \includegraphics[width=1\linewidth]{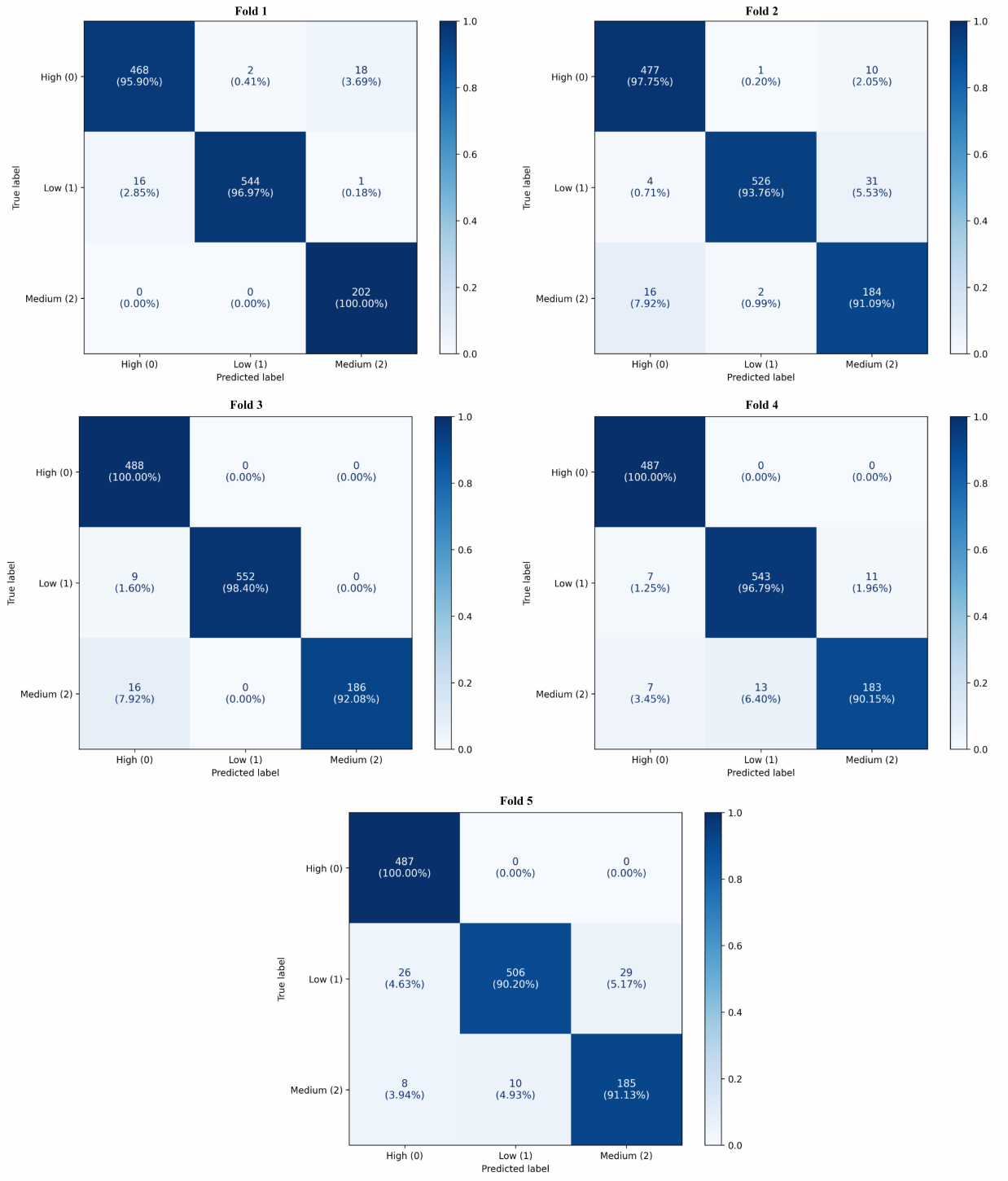}
    \caption{Confusion matrices representing the classification performance of the DualEngage architecture across five-fold cross-validation for group engagement recognition.}\label{fig:cm}
\end{figure}

Our proposed network attains an overall accuracy of $0.9621 \pm 0.0161$ with a macro-averaged F1 of $0.9530 {\pm}0.0204$ across five-fold validation. Per-class performance can be found in figure \ref{fig:cm} showing the confusion matrices for all folds. 
\begin{table}
\caption{Five-fold results on group engagement level recognition. (\emph{Bold} = best per column.)}
\label{tab:fivefold}
\centering
\begin{tabular}{lcccc}
\toprule
\textbf{Fold} & \textbf{Accuracy} & \textbf{Macro Precision} & \textbf{Macro Recall} & \textbf{Macro F1 Score} \\
\midrule
Fold 1 & 0.9704 & 0.9591 & \textbf{0.9762} & 0.9670 \\
Fold 2 & 0.9488 & 0.9240 & 0.9420 & 0.9318 \\
Fold 3 &\textbf{0.98} & \textbf{0.9838} & 0.9682 & \textbf{0.9752} \\
Fold 4 & 0.9696 & 0.9640 & 0.9565 & 0.96 \\
Fold 5 & 0.9416 & 0.9266 & 0.9378 & 0.9311 \\
\midrule
\textbf{Mean} $\pm$ \textbf{Std} 
& $0.9621 \pm 0.0161$ 
& $0.9515 \pm 0.0257$ 
& $0.9561 \pm 0.0165$ 
& $0.9530 \pm 0.0204$ \\
\bottomrule
\end{tabular}
\end{table}
The model demonstrates consistent performance across all folds, with 
narrow standard deviations in both accuracy and macro-F1, indicating robustness to data split variability. The high macro-recall of $0.9561 \pm 0.0165$ suggests that the model recovers well across all three engagement classes, including the minority \textit{Medium} class. We attribute this improvement to two architectural contributions: attention pooling, which focuses the motion representation on the most behaviorally salient students rather than averaging across all tracked individuals, and softmax-gated fusion, which adaptively adjusts the relative contribution of the motion and scene streams on a per-sample basis rather than assuming equal importance.

\begin{figure}[htbp]
    \centering
    \includegraphics[width=1\linewidth]{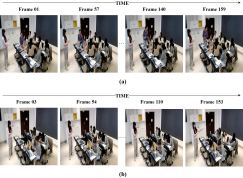}
    \caption{Representative frames sampled across classroom clips in a chessboard layout. (a) High engagement example: students maintain attention toward the instructor/board, showing consistent orientation and participation. (b) Low engagement example: several students exhibit reduced attention, such as head-down posture or looking away, indicating weaker collective involvement.}
    \label{fig:misdetect}
\end{figure}

Residual confusions between \textit{Low} and \textit{High} likely arise from clips exhibiting minimal per-student motion but strong scene-level synchrony, a setting where the local and global cues partially disagree. As illustrated in Figure \ref{fig:misdetect}, a high-engagement group may remain physically still, especially in a chessboard layout, producing weak optical-flow magnitude. Conversely, clips with low engagement are generally also low-motion. In such regimes, the primary stream becomes less discriminative, and the prediction relies more heavily on global context, leading to confusions between the \textit{High} and \textit{Low} class. Future work may incorporate attention to teacher actions and fine-grained head or gaze cues to further disambiguate such cases.
\subsection{Ablation Study}
\label{sec:ablation}
We conduct an ablation study to quantify the contribution of each component in DualEngage, reporting accuracy and macro-averaged precision, recall, and F1-score (Table~\ref{tab:single_branch_performance}). The primary stream, when used in isolation, performs poorly due to several challenges. One major reason is variations in lighting, occlusions, and subtle or inconsistent movements, which can cause the flow estimates to become unreliable, leading to imprecise motion representations. In scenarios with heavy crowding or occlusion, individual students may be temporarily lost from the frame or mistakenly tracked under incorrect IDs, causing gaps or fragmentation in their motion data. Similarly, the secondary stream alone also shows suboptimal performance, despite considering group-level context. While it captures spatial-temporal effectively, it lacks the detailed information of individual students provided by the primary stream. To assess the importance of temporal modeling, we remove the transformer encoder from the fused architecture. This variant drops sharply in macro-F1  despite a relatively high macro-precision, suggesting a conservative prediction behavior where fewer classes are predicted confidently, inflating precision while reducing recall and overall class-balanced performance. These results confirm that temporal encoding is critical for stabilizing class-wise predictions and capturing engagement dynamics over time. Similarly, using an LSTM block or 1D CNN block instead of a transformer encoder prove to be inefficient, as seen from the table. In contrast, the full DualEngage model outperforms both single-branch models by a large margin, achieving the highest accuracy, precision, and F1-score.

\begin{table}
\caption{Ablation study of DualEngage comparing the full model with single-stream baselines, temporal modeling removal, and alternative backbone choices.)}
\label{tab:single_branch_performance}
\centering
\begin{tabular}{lcccc}
\toprule
\textbf{Fold} & \textbf{Accuracy} & \textbf{Macro Precision} & \textbf{Macro Recall} & \textbf{Macro F1 Score} \\
\midrule
Primary Stream Only & 0.5321 & 0.5257 & 0.4920 & 0.5083 \\
Secondary Stream Only & 0.6214 & 0.6075 & 0.5894  & 0.5983 \\
DualEngage w/o Temporal Encoder & 0.5167 & 0.5257 &  0.5167 & 0.5043 \\
DualEngage LSTM & 0.6017 & 0.7157 &  0.6017 & 0.5747 \\
DualEngage 1DCNN & 0.4950 & 0.5066 &  0.4950 & 0.4949 \\
DualEngage + R(2+1)D backbone & 0.9012 & 0.9145 & 0.8824 & 0.8982 \\
DualEngage + I3D backbone     & 0.8945 & 0.9021 & 0.8910 & 0.8965 \\
DualEngage w/o attention pooling \& gated fusion& 0.8823 & 0.8866 & 0.8634  & 0.8706 \\
DualEngage (Proposed) & \textbf{0.9621} & \textbf{0.9515} &  \textbf{0.9561}& \textbf{0.9530} \\

\bottomrule
\end{tabular}
\end{table}

We also evaluate DualEngage excluding the attention pooling in \ref{attn_pool} and utilizing simple concatenation instead of adaptive gated fusion of the two branches. This further proves the significance of adding these modules within the framework. Finally, we evaluate alternative backbones within the dual-stream framework. DualEngage with R(2+1)D achieves the strongest macro-precision and a competitive macro-F1, while the full DualEngage configuration attains the best overall accuracy  and macro-recall, precision and f1 score, reflecting a better balance across engagement classes. Overall, the ablations support our hypothesis that combining person-centric motion cues with global scene context, together with explicit temporal modeling, is essential for reliable group engagement recognition.

\section{Conclusion} \label{sec: conclusion}
This paper presents DualEngage, a dual-stream architecture for group-level 
student engagement recognition from in-classroom videos. The central premise 
of this work is that group engagement relies on both individual motion cues and global scene context. To operationalize this, DualEngage combines a transformer-encoded motion stream that models the temporal evolution of per-student optical flow with a 3D ResNet-18 scene stream that captures spatiotemporal context from the full video clip. Beyond the dual-stream design itself, two architectural contributions address 
fundamental limitations of naive feature aggregation. First, attention pooling over per-student transformer embeddings replaces uniform mean pooling with a learned, importance-weighted aggregation, allowing the model to emphasize students exhibiting stronger behavioral signals. Second, 
softmax-gated fusion replaces static concatenation with an input-conditioned 
convex combination of the two stream representations, enabling the model to 
dynamically shift reliance between motion and scene cues on a per-sample basis. Together, these mechanisms adjust to the varying informativeness of individual and collective cues. In a stratified five-fold evaluation on the OUC–CGE dataset, the proposed dual-stream model achieved stable performance and substantially outperformed both single-stream variants, supporting our hypothesis. The ablation study confirms that each component contributes independently to the overall performance and that their combination is necessary to achieve reliable classification.

Despite its success, the architecture introduces notable computational cost due to dense optical flow and per-student motion tube extraction. RAFT inference is more expensive than lightweight flow models and requires GPU acceleration for practical training time. Additionally, the pipeline depends on accurate person detection and tracking; long-term occlusion, heavy crowding, or missed detections can lead to incomplete motion tubes. Future work may address low-motion ambiguity by incorporating gaze, posture, or teacher cues.
\\

\noindent \textbf{Conflict of Interest} \\
\noindent No conflict of interest to declare 
\\

\noindent \textbf{Data Availability Statement} \\
\noindent The dataset used in this study can be made available with a reasonable request to the corresponding author. 
\\ 

\noindent \textbf{Funding} \\
\noindent This study did not receive any funding.
\\











\bibliographystyle{cas-model2-names}

\bibliography{cas-refs}



\end{document}